\documentclass{article}
\usepackage{spconf,amsmath,graphicx}

\newcommand{\etal}{\emph{et al.}}
\usepackage{graphicx}
\usepackage{amsmath,amssymb} % define this before the line numbering.
\usepackage{mathtools}
\usepackage{color}
\usepackage{booktabs}
\usepackage{algorithm}
\usepackage{algpseudocode}
\usepackage[export]{adjustbox} % Vertical alignment
\usepackage{array,multirow}
\usepackage{url}
\usepackage{tikz}

\DeclarePairedDelimiter\norm{\lVert}{\rVert}

\title{Learning Robust Self-Attention Features For Speech Emotion Recognition with Label-Adaptive Mixup}
\name{Lei Kang$^{\dag}$, Lichao Zhang$^{\ddag}$, Dazhi Jiang$^{\dag}$\thanks{This work has been partially supported by the grants 62206163 and 62006245 from National Natural Science Foundation of China, the grant 140/09421059 from Shantou University, and STU Incubation Project for the Research of Digital Humanities and New Liberal Arts. }}
\address{$^{\dag}$Computer Science Dept., Shantou University, China\\
$^{\ddag}$Aeronautics Engineering College, Air Force Engineering University, China\\
\tt \small \{lkang, dzjiang\}@stu.edu.cn, lichao.zhang@outlook.com}

\begin{document}
%\ninept
%
\maketitle
\begin{abstract}
Speech Emotion Recognition (SER) is to recognize human emotions in a natural verbal interaction scenario with machines, which is considered as a challenging problem due to the ambiguous human emotions. Despite the recent progress in SER, state-of-the-art models struggle to achieve a satisfactory performance. We propose a self-attention based method with combined use of label-adaptive mixup and center loss. By adapting label probabilities in mixup and fitting center loss to the mixup training scheme, our proposed method achieves a superior performance to the state-of-the-art methods.
\end{abstract}


\section{Introduction}
\label{sec:intro}

Speech Emotion Recognition (SER) is one of the most important research topics in the field of human-computer interaction. SER tries to classify input speech signals into their corresponding emotion categories, which is a challenging problem because of the inherent complexity, ambiguousness, and high personality of human emotions. How to extract the emotional features effectively is the key to solve SER problems. 

Recently, deep neural network (DNN) based methods have dominated the field of SER. Especially with the success of convolutional neural network (CNN) in computer vision domain, researchers usually transform speech signals into hand-crafted spectrogram features as input so as to take advantage of the CNN models~\cite{zhong2020lightweight,wu2021emotion,wu2022neural,chernykh2017emotion,xu2020improve}. But the raw speech waveforms can also be utilized directly as input thanks to the development of recurrent neural network (RNN)~\cite{sarma2018emotion}. However, RNN-based models always struggle with vanishing gradient problem for long speech signals. 

Self-attention mechanism has attracted significant attention in the speech processing community~\cite{tarantino2019self,liu2020speech}. More recently, excellent self-supervised models have emerged, of which wav2vec2.0~\cite{wav2vec2} and HuBERT~\cite{hubert} are ones of the most popular and performant models. Furthermore, a bunch of pre-trained models of wav2vec2.0 and HuBERT are available, which have already initialized a good weight distribution for general purpose in the speech domain. We take HuBERT as our baseline architecture and adapt it to SER with some essential modifications.

To further improve generalization capability of SER model, data augmentation techniques are widely used, among which mixup strategy is proved to be a simple and effective method by mixing pairs of training data and their labels~\cite{zhu2022speech}. Dai~\etal~\cite{dai2019learning} proposed a SER method with learning objectives of both center loss and recognition loss. The center loss pulls features in the same class closer to their class center while the recognition loss separates features from different emotional categories. However, the combined use of both mixup and center loss has not been reported, because mixup generates mixed labels with probabilities while center loss asks for class indexes. We propose an effective method to use both mixup and center loss towards achieving a better performance on SER tasks by learning robust emotional features.

Our main contributions are threefold: firstly, we modify a HuBERT-based self-attention model to extract emotional features in a more effective way, which is illustrated by a comprehensive ablation study. Secondly, we propose a label-adaptive mixup method boosting SER performance significantly. And thirdly, to the best of our knowledge, it is the first attempt for combining center loss and mixup together to SER. Our proposed method achieves a superior performance to the state of the arts on IEMOCAP speech dataset with $75.37\%$ WA and $76.04\%$ UA in Leave-One-Session-Out (LOSO) fashion. Our code is available at \url{https://github.com/leitro/LabelAdaptiveMixup-SER}.

%%%%%%%%%%%%%%%%%%%%%%%%%%%%%%%%%%%%%%%%%%%%%%%%%%%%%%
\section{Speech Emotion Recognition}
\begin{figure}[h!]
    \centering
    \includegraphics[width=0.9\linewidth, height=10.5cm]{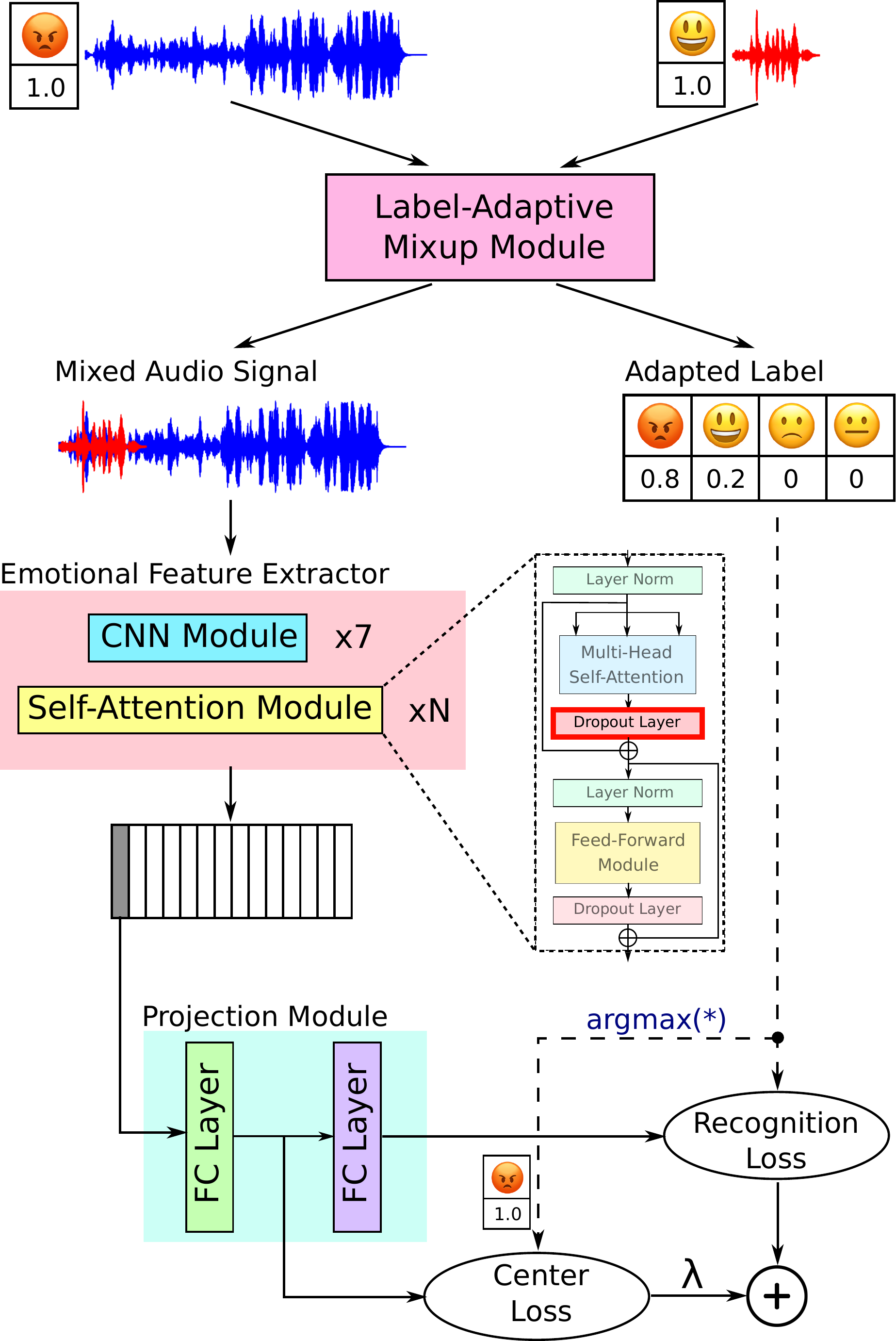}
    \caption{Illustration of our proposed SER model.}
    \label{fig:arch}
    \vspace{-0.4cm}
\end{figure}

In this section, we propose our SER model as shown in Figure~\ref{fig:arch}, which consists of 3 main parts: label-adaptive mixup module, emotional feature extractor and projection module. Let $\{\mathcal{X}, \mathcal{Y}\}$ be an emotional speech dataset, containing speech signals $x \in \mathcal{X}$ and their corresponding one-hot encoded emotion categories $y \in \mathcal{Y}$. $E$ refers to the emotion categories as angry, happy, sad and neutral.

\subsection{Label-Adaptive Mixup}
Mixup~\cite{zhang2018mixup} is a popular data-agnostic data augmentation technique that trains a neural network on convex combinations of pairs of examples and their labels. Given random training pairs $(x_i, y_i)$ and $(x_j, y_j)$, we can obtain a pair of synthetic example $(x_{ij}, y_{ij})$ by the conventional mixup strategy as follows:

\vspace{-0.5cm}
\begin{eqnarray}
    x_{ij} &=& \lambda x_i + (1-\lambda) x_j\\
    \label{equ:mixupx}
    y_{ij} &=& \lambda y_i + (1-\lambda) y_j
    \label{equ:mixupy}
\end{eqnarray}
where $\lambda \sim \mathcal{B}(\alpha, \alpha) \in [0, 1]$ and $\mathcal{B}$ refers to Beta distribution with $\alpha \in (0, \infty)$. Thus, mixup is a straightforward method to augment training data by applying linear interpolation in the feature space.

The speech data has variable length according to its textual content, but its label is an emotional category with probability of $1$. Thus, it is less accurate to treat the labels as same as the speech clips as shown in Equation~\ref{equ:mixupy}. We propose our label-adaptive mixup method to replace it as follows:

\begin{equation}
	y_{ij} = \left(\frac{\lambda l_i}{\lambda l_i+(1-\lambda) l_j}\right) y_i + \left(\frac{(1-\lambda) l_j}{\lambda l_i+(1-\lambda) l_j}\right) y_j
\end{equation}

where $y_{ij}$ is a list of emotion categories $[z_1, z_2, ..., z_{|E|}]$ summing up to $1$ and $l_i$ is the length of $i$-th sample. To put it simple, we assign $\lambda$ to be a constant $0.5$. Thus, the probabilities of emotion categories depend only on the lengths of the input speech data pair.

\subsection{Emotional Feature Extraction}

Emotional feature extractor and projection module constitute the pipeline of effective emotional feature extraction. We choose the latest release of Hidden Unit BERT (HuBERT)~\cite{hubert} as our baseline model for emotional feature extractor. There are 3 architectures of HuBERT, which are HuBERT-Base, HuBERT-Large and HuBERT-XLarge. HuBERT-Large is chosen as our baseline model, which is pre-trained on 60,000 hours of unlabeled audio from Libri-Light dataset~\cite{kahn2020}. HuBERT-Large model consists of a convolutional part and a Transformer part. We keep the convolutional part unchanged and focus on tuning the latter one for SER tasks. The Transformer part consists of 24 self-attention modules as shown in the dashed rectangle in Figure~\ref{fig:arch}. We reduce the number of self-attention modules and modify the dropout probability between multi-head self-attention and feed-forward module as highlighted in red rectangle. We will discuss these modifications later in Section~\ref{sec:ablation}.

% \begin{figure}[h!]
%     \centering
%     \includegraphics[width=0.28\linewidth]{images/selfattn.pdf}
%     \caption{Self-attention module in the Transformer part of HuBERT-Large model.}
%     \label{fig:selfattn}
% \end{figure}

We feed speech data $x \in \mathcal{X}$ into the emotional feature extractor and the high-level emotional feature representation $F_e$ is produced. $F_e$ is a sequence of feature vectors with variable length according to different input length of speech signals. Instead of using average pooling~\cite{morais2022speech} to aggregate the sequence of feature vectors into fixed-size, we simply take the first feature vector $F_{e}^{0}$ as the emotional feature representation for the whole sequence, thanks to the great capability of long-range feature exploring and extraction of self-attention modules. We will compare it with average pooling method in Section~\ref{sec:ablation}. Then, as shown in the bottom of Figure~\ref{fig:arch}, two fully-connected layers are stacked in the projection module, which are denoted as $f_0$ and $f_1$ for the first (green) and second(purple) layer, respectively.

\subsection{Learning Objectives}

\subsubsection{Recognition Loss}
Log-softmax Kullback-Leibler divergence loss is utilized as our recognition loss to guide the SER model for emotion classification, which is presented as follows:

\begin{equation}
    \mathcal{L}_r = \sum_{k=1}^{|E|} z_k \log\left(\frac{z_k}{\hat{z}_k}\right)
    \label{equ:rec}
\end{equation}

where $z_k$ is the groundtruth probability of $k$-th emotion category in $y_{ij}$, and $\hat{z}_k$ is the predicted probability for $k$-th emotion in $E$. $\hat{z}_k \in \hat{y}_{ij}$, which is obtained by applying Softmax on the output feature $f_1(f_0(F_{e}^{0}))$.

\subsubsection{Center Loss}
\label{sec:centerloss}

Center loss was first proposed and utilized for face recognition~\cite{centerloss}. It updates feature centers of training data per mini-batch and tries to reduce the intra-class variations on the feature space. Dai~\etal~\cite{dai2019learning} have applied center loss for illustrating its capability to learn more effective features for SER tasks. To work with mixup strategy during training, we modify the formula of center loss as follows:

\begin{equation}
    \mathcal{L}_c = \frac{1}{N} \sum_{i=1}^{N} \norm{f_0(F_{e}^{0}) - \mu_{argmax(y_{ij})}}_{2}^{2}
    \label{equ:center}
\end{equation}

where $N$ is the number of training samples in a mini-batch, and $\mu_{argmax(y_{ij})}$ is the feature centroid for emotion category $argmax(y_{ij})$. $y_{ij}$ is a list of probabilities on emotion categories $E$ with the usage of mixup method, and only the emotion category with the highest probability is selected as groundtruth for center loss. In this way, not only we solve the problem that mixup and center loss didn't use to work together, but also robust emotional features could be learned by introducing mixed noise. Thus, the model is trained using a joint loss as follows:

\begin{equation}
    \mathcal{L} = \mathcal{L}_r + \lambda \mathcal{L}_c
\end{equation}

where $\lambda$ is a trade-off hyper-parameter for balancing both of the losses.

%%%%%%%%%%%%%%%%%%%%%%%%%%%%%%%%%%%%%%%%%%%%%%%%%%%%%%
\section{Experiments}

\subsection{Dataset and Metrics}

The IEMOCAP~\cite{Busso2008} dataset is utilized to evaluate our method. It consists of approximately 12 hours of multimodal data with speech, transcriptions and facial recordings. We only focus on the speech data in this work. There are 5 sessions in the speech data, in each of which a conversation between 2 exclusive speakers is involved. To make our results comparable to the state-of-the-art works~\cite{wu2021emotion,wu2022neural,zou2022speech}, we merge "excited" into "happy" category and use speech data from four categories of "angry", "happy", "sad" and "neutral", which leads to a 5531 acoustic utterances in total from 5 sessions and 10 speakers.

The widely used Leave-One-Session-Out (LOSO) 5-fold cross-validation is utilized to report our final results. Thus, at each fold, 8 speakers in 4 sessions are used for training while the other 2 speakers in 1 session are used for testing. Both the Weighted Accuracy (WA) and Unweighted Accuracy (UA) are chosen as the evaluation metrics.

\subsection{Implementation Details}

For the optimization, the model is trained using Adam algorithm with a dynamic learning rate scheme (reducing by a factor of $1.25$ at each epoch until 20th epoch) for both recognition loss and center loss. The learning rates are initialized as $1e\text{-}4$ and $1e\text{-}3$ for recognition loss and center loss, respectively. All the experiments are done on a NVIDIA RTX3090. The model is implemented with PyTorch 1.12, and please refer to our code for more details.

% For the optimization, the model is trained using Adam algorithm ($\beta_1=0.9$, $\beta_2=0.98$ and $\epsilon=1e-9$) with a dynamic learning rate scheme (reducing by a factor of $1.25$ at each epoch until 20th epoch) for both recognition loss and center loss. The learning rates are initialized as $1e-4$ and $1e-3$ for recognition loss and center loss, respectively. All the experiments are done on a NVIDIA RTX3090. The model is implemented with PyTorch 1.12, and please refer to our code for more details.

\subsection{Baseline Model}
\label{sec:ablation}

We try to explore the best use of HuBERT-Large model for the SER tasks. In this section, all the experiments are done with exact 5 epochs training on the speech data of first 8 speakers in 4 sessions, and the WA and UA results are reported by evaluating on the remaining 2 speakers in the 5th session. In this way, we can not only ensure the speaker-independent setting in the experiments, but also conduct the experiments effectively without seeking for the best epoch. 

\begin{figure}[h!]
    \centering
    \includegraphics[width=0.98\linewidth]{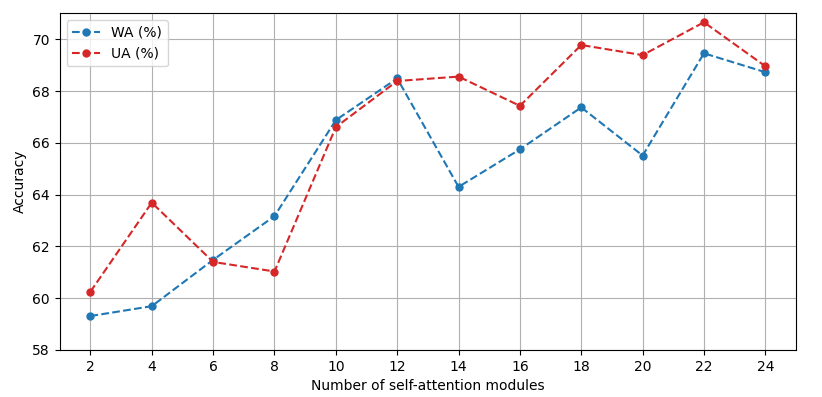}
    \caption{Ablation study curves according to the number of self-attention modules to use. }
    \label{fig:layer}
    \vspace{-0.4cm}
\end{figure}

Firstly, as HuBERT-Large model is huge with 24 self-attention modules, we want to know how the SER performance relates to the number of self-attention modules. From Figure~\ref{fig:layer}, the best performance is achieved with the usage of $22$ self-attention modules. We can also see that the performance is not always getting better with more layers, $12$ is also a good number to choose with a balance of performance and efficiency. But as our goal in this paper is to exploit the best performance of the proposed method, $22$ is the final selection.

% \begin{table}[t!h]
%     \caption{Selection of top N self-attention modules (24 in total) in the transformer part of HuBERT-Large model.}
%     \label{tab:layer}
%     \centering
%     \small
%     \begin{tabular}{ccc}
%     \toprule
%     \textbf{$N$ Self-Attn Modules} & \textbf{WA (\%)} & \textbf{UA(\%)}\\
%     \midrule
%     2 & 59.31 & 60.25 \\
%     4 & 59.69 & 63.69 \\
%     6 & 61.48 & 61.40 \\
%     8 & 63.17 & 61.03 \\
%     10 & 66.88 & 66.61 \\
%     12 & 68.49 & 68.39 \\
%     14 & 64.30 & 68.56 \\
%     16 & 65.75 & 67.43 \\
%     18 & 67.37 & 69.78 \\
%     20 & 65.51 & 69.39 \\
%     \textbf{22} & \textbf{69.46} & \textbf{70.66} \\
%     24 & 68.73 & 68.96 \\
%     \bottomrule
%     \end{tabular}
% \end{table}

\begin{table}[t!h]
    \caption{Dropout probability of the projection dropout layer between multi-head self-attention and feed-forward module.}
    \vspace{0.1cm}
    \label{tab:dropout}
    \centering
    \small
    \scalebox{0.88}{
    \begin{tabular}{ccc}
    \toprule
    \textbf{Dropout Prob.} & \textbf{WA (\%)} & \textbf{UA(\%)}\\
    \midrule
		0 & 69.46 & 70.66\\
		0.1 & 69.94 & 70.49 \\
		0.2 & 69.46 & 70.41 \\
		0.3 & 70.59 & 70.61 \\
		\textbf{0.4} & \textbf{70.99} & \textbf{72.83} \\
		0.5 & 61.97 & 67.35 \\
    \bottomrule
    \end{tabular}
    }
    \vspace{-0.2cm}
\end{table}

Secondly, zooming into a self-attention module as visualized in the dashed rectangle of Figure~\ref{fig:arch}, the multi-head self-attention extracts the contextual information among the sequential speech features, while the feed-forward module tries to obtain high-level emotional features. Thus, the projection dropout layer in between plays the key role and need to be adjusted so as to prevent over-fitting towards a specific task. According to Table~\ref{tab:dropout}, we choose $0.4$ for the projection dropout layer at each self-attention module in the emotional feature extractor.

\subsection{Ablation Study}

Based on the previous section, we have find the best architecture for HuBERT-Large model as the emotional feature extractor. In this section, we further discuss feature reduction methods, mixup methods and the use of center loss. For the experiments, we still train the model on the first 4 session data and report the WA and UA results by evaluating on the remaining session. But we randomly fetch out $10\%$ of training data as a validation set, on which $10$-epoch early stopping strategy is applied to find the best model weights. Then the WA and UA results can be obtained by evaluating the best model on the test data.

As shown in Figure~\ref{fig:arch}, the emotional feature $F_e$, i.e. the output of the emotional feature extractor, is a variable-length sequence of vectors, which need to be summarized into a fixed-size vector for the projection module. Here we compare two simple ways: down-sampling with adaptive average pooling, namely $Avg(F_e)$, or simply selecting the first vector of $F_e$, namely $F_{e}^{0}$. The latter achieves a better performance according to the results as shown in the first 2 rows of Table~\ref{tab:abla}. It is because that the related emotional feature has been aggregated into this single vector during training, which is more robust and reliable than the hand-crafted pooling one.

% \begin{table}[t!h]
%     \caption{Comparison of ways to transform variable-length feature into fixed-size.}
%     \vspace{0.1cm}
%     \label{tab:avgpool}
%     \centering
%     \small
%     \begin{tabular}{ccc}
%     \toprule
%     \textbf{Idea} & \textbf{WA (\%)} & \textbf{UA(\%)}\\
%     \midrule
%         $AdaptAvgPool(F_e)$ & 70.91 & 71.80\\
%         \textbf{First Vector Selection $F_{e}^{0}$} & \textbf{70.99} & \textbf{72.83}\\
%     \bottomrule
%     \end{tabular}
%     \vspace{-0.5cm}
% \end{table}

%\subsection{Evaluation on Proposed Modules}

\begin{table}[t!h]
 %\caption{Ablation study on our proposed robust self-Attention model by analysing on the effectiveness of different components, including the ways to transform variable-length feature into fixed-size, two different mixup methods, i.e. conventional and label adaption, and center loss. Additionally, we use trade off hyper-parameter $\lambda$ for weighting the center loss.}
    \caption{Ablation study on our proposed methods including Feature Reduction method, Mixup method and Center Loss method, from left to right respectively.}
    \vspace{0.1cm}
    \label{tab:abla}
    \centering
    %\color{blue}
    \small
    \scalebox{0.86}{
    \begin{tabular}{cc|cc|c|cc}
    \toprule
    \multicolumn{2}{c|}{\textbf{Feat. Reduct.}} & \multicolumn{2}{c|}{\textbf{Mixup}} &\textbf{Center Loss} & \multirow{2}{*}{\textbf{WA (\%)}} & \multirow{2}{*}{\textbf{UA (\%)}}\\ 
    $Avg(F_e)$ & \textbf{$F_{e}^{0}$}&Conv. &{Adapt.}& \textbf{$\lambda$} &  & \\ \hline
    %\midrule
    \checkmark & $-$ &$-$ & $-$ & 0 & 70.91 & 71.80 \\
    $-$ & \checkmark &$-$ & $-$ & 0 & 70.99 & 72.83 \\
    $-$ & \checkmark &\checkmark & $-$ & 0 & 70.83 & 74.06 \\
    $-$ & \checkmark &$-$ & \checkmark & 0 & 73.97 & 75.03 \\
    $-$ & \checkmark &$-$ & \checkmark & 0.0005 & 74.54 & 76.20\\
    $-$ & \checkmark &$-$ & \checkmark & 0.001 & 74.21 & 75.99\\
    $-$ & \checkmark &$-$ & \checkmark & \textbf{0.002} & \textbf{74.86} & \textbf{76.31}\\	
    \bottomrule
    \end{tabular}
    }
    \vspace{-0.2cm}
\end{table}

To evaluate the effectiveness of our proposed label-adaptive mixup method, we make use of the conventional mixup~\cite{zhang2018mixup} method as comparison. Since mixup is considered as one of the data augmentation techniques, we also adopt some common data augmentation techniques together with mixup for the following experiments such as Gaussian Noise, Clipping Distortion, Gain, Gain Transition, Polarity Inversion, Tanh Distortion, Time Mask, Time Stretch and Pitch Shift. With the random combination of these common data augmentation techniques and the use of conventional mixup method, the SER model achieves $70.83\%$ and $74.06\%$ for WA and UA, respectively, as shown in the 3rd row of Table~\ref{tab:abla}. Compared with the conventional mixup strategy, our proposed label-adaptive mixup method boost the performance by approximately $3\%$ on WA and $1\%$ on UA as shown in the 4th row of Table~\ref{tab:abla}. Such a huge boost is obtained because the proposed method re-balance the weights of emotional categories according to the variable lengths of speech clips. In the common cases especially from the IEMOCAP dataset, a single emotional category is consistent in either a short interjection or a long monologue, such that the conventional mixup would introduce strong noise by treating both interjection and monologue equally. 

% \begin{table}[t!h]
%     \caption{Data augmentation techniques.}
%     \label{tab:augment}
%     \centering
%     \small
%     \begin{tabular}{lcc}
%     \toprule
%     \textbf{Technique} & \textbf{WA (\%)} & \textbf{UA(\%)}\\
%     \midrule
%     	$-$ & 70.99 & 72.83 \\
%     \midrule
% 		\circled{1} Gaussian Noise & 68.73 & 70.43 \\
% 		\circled{2} Clipping Distortion & 70.51 & 73.82 \\
% 		\circled{3} Gain & 68.33 & 72.56 \\
% 		\circled{4} Gain Transition & 70.59 & 72.26 \\
% 		\circled{5} Polarity Inversion & 70.67 & 72.19 \\
% 		\circled{6} Tanh Distortion & 69.38 & 70.84 \\
% 		\circled{7} Time Mask & 69.22 & 70.03 \\
% 		\circled{8} Time Stretch & 67.45 & 69.70 \\
% 		\circled{9} Pitch Shift & 50.93 & 59.63 \\
% 	\midrule
% 		\circled{2} + \circled{3} & 66.32 & 68.93 \\
% 		\circled{2} + \circled{4} & 66.40 & 69.77 \\
% 		\circled{2} + \circled{5} & 70.75 & 73.45 \\
% 		\circled{3} + \circled{4} & 70.59 & 71.41 \\
% 		\circled{3} + \circled{5} & 70.83 & \textbf{74.06} \\
% 		\circled{4} + \circled{5} & \textbf{71.88} & 73.76 \\
%     \bottomrule
%     \end{tabular}
% \end{table}

Furthermore, we try to equip a center loss in the training phase. As explained in Section~\ref{sec:centerloss}, $\lambda$ is a hyper-parameter to trade off center loss against recognition loss. From 5 to 7th row of Table~\ref{tab:abla}, we demonstrate the effect on performance with different $\lambda$. The best performance is achieved at $\lambda = 0.002$. 

% \begin{table}[t!h]
%     \caption{Comparison between conventional and label-adaptive mixup methods.}
%     \vspace{0.1cm}
%     \label{tab:mixup}
%     \centering
%     \small
%     \begin{tabular}{ccc}
%     \toprule
%     \textbf{Label Adaption} & \textbf{WA (\%)} & \textbf{UA(\%)}\\
%     \midrule
%     	$-$ & 70.83 & 74.06\\
% 		\textbf{\checkmark} & \textbf{73.97} & \textbf{75.03}\\
%     \bottomrule
%     \end{tabular}
%     \vspace{-0.2cm}
% \end{table}

% \begin{table}[t!h]
%     \caption{Comparison w/ and w/o center loss. $\lambda$ is the hyper-parameter to trade off center loss against recognition loss.}
%     \vspace{0.1cm}
%     \label{tab:centerloss}
%     \centering
%     \small
%     \begin{tabular}{cccc}
%     \toprule
%     \textbf{Center Loss} & \textbf{$\lambda$} & \textbf{WA (\%)} & \textbf{UA(\%)}\\
%     \midrule
%     	$-$ & $-$ & 73.97 & 75.03 \\
%     	\checkmark & 0.0005 & 74.54 & 76.20\\
%     	\checkmark & 0.001 & 74.21 & 75.99\\
%     	\textbf{\checkmark} & \textbf{0.002} & \textbf{74.86} & \textbf{76.31}\\	
%     \bottomrule
%     \end{tabular}
%     \vspace{-0.2cm}
% \end{table}

\subsection{Comparison with State Of The Arts}

Finally, we have found the best neural network architecture and hyper-parameters for SER according to the evaluation results on the data of last $2$ speakers of $5$-th session in IEMOCAP, which is only one fold. So we do the full 5-fold cross-validation in LOSO fashion and report the average results on WA and UA as shown in Table~\ref{tab:sota}, achieving a superior performance among state of the arts.

% \begin{table}[t!h]
%     \caption{Leave-one-session-out (LOSO) 5-fold cross validation.}
%     \label{tab:loso_fin}
%     \centering
%     \small
%     \begin{tabular}{ccc}
%     \toprule
%     \textbf{Fold} & \textbf{WA (\%)} & \textbf{UA(\%)}\\
%     \midrule
%     	1 & 74.86 & 76.31\\
%     	2 & 79.18 & 80.89\\
%     	3 & 74.54 & 74.70\\
%     	4 & 73.71 & 72.36\\
%     	5 & 74.56 & 75.93\\
%     \midrule
%         \textbf{Mean} & \textbf{75.37} & \textbf{76.04}\\
%     \bottomrule
%     \end{tabular}
% \end{table}

% \begin{table}[t!h]
%     \caption{Random split 5-fold cross-validation.}
%     \label{tab:rand_fin}
%     \centering
%     \small
%     \begin{tabular}{ccc}
%     \toprule
%     \textbf{Fold} & \textbf{WA (\%)} & \textbf{UA(\%)}\\
%     \midrule
%     	1 & 77.49 & 78.42\\
%     	2 & 80.20 & 81.18\\
%     	3 & 79.75 & 80.74\\
%     	4 & 78.57 & 79.18\\
%     	5 & 79.67 & 80.56\\
%     \midrule
%         \textbf{Mean} & \textbf{79.14} & \textbf{80.02}\\
%     \bottomrule
%     \end{tabular}
% \end{table}

%%% leave-one-session-out 5 fold cross validation results
\begin{table}[t!h]
    \caption{Comparison with state of the arts by Leave-One-Session-Out (LOSO) 5-fold cross-validation.}
    \vspace{0.1cm}
    \label{tab:sota}
    \centering
    \small
    \scalebox{0.89}{
    \begin{tabular}{cccc}
    \toprule
    \textbf{Method} & \textbf{Year} & \textbf{WA (\%)} & \textbf{UA(\%)}\\
    \midrule
    Human Performance~\cite{chernykh2017emotion}  & 2017 & 69.00 & 70.00\\
    TDNN-LSTM-attn~\etal~\cite{sarma2018emotion} & 2018 & 70.10 & 60.70\\
    LSTM~\etal~\cite{parry2019analysis} & 2019 &56.99 & 53.07\\
    IS09-classification~\etal~\cite{tarantino2019self} & 2019 & 64.33 & 64.79\\
    CNN-GRU-SeqCap~\etal~\cite{wu2019speech} & 2019 & 72.73 & 59.71\\
    HGFM~\etal~\cite{xu2020hgfm} & 2020 & 66.60 & 70.50\\
    ACNN~\etal~\cite{xu2020improve} & 2020 & 67.28 & 67.94\\
    ASR-SER~\etal~\cite{feng2020end} & 2020 & 68.60 & 69.70\\
    Lightweight model~\etal~\cite{zhong2020lightweight} & 2020 & 70.39 & 71.72\\
    SSL\&CMKT fusion~\etal~\cite{zhang2021combining} & 2021 & 61.16 & 62.50\\
    Audio$_{25,250}$+BERT~\etal~\cite{wu2021emotion} & 2021 & 69.44 & 70.90\\
    Selective MTL~\etal~\cite{zhang2022selective} & 2022 & 56.87 & 59.47\\
    MFCC+Spectrogram+W2E~\etal~\cite{zou2022speech} & 2022 & 69.80 & 71.05\\
    CNN-SeqCap~\etal~\cite{wu2022neural} & 2022 & 70.54 & 56.94\\
    \midrule
    \textbf{Proposed} & \textbf{2023} & \textbf{75.37} & \textbf{76.04}\\
    \bottomrule
    \end{tabular}
    }
    \vspace{-0.2cm}
\end{table}

\section{Conclusion}

In this paper, we present a self-attention based SER method, whose architecture and hyper-parameters have been modified and evaluated in depth. Furthermore, we propose a simple and effective label-adaptive mixup method, which boosts the performance drastically. Finally, as far as we know, we are the first to train a SER model with combined use of mixup and center loss, which forces the model to learn more robust features. Comparing with the state-of-the-art works, our proposed method has achieved a superior performance on IEMOCAP speech dataset.

\vfill\pagebreak

%\section{REFERENCES}
%\label{sec:refs}

\small
\bibliographystyle{IEEEbib}
\bibliography{refs}

\end{document}